\documentclass[]{elsarticle}

\usepackage{amssymb}

\usepackage{graphicx, inputenc,url,graphicx,booktabs, float,tcolorbox,amsmath,vcell,colortbl, multirow, subcaption, multicol, setspace, upgreek,color }

\usepackage{siunitx}

\usepackage{amsmath,amssymb,amsfonts}%
\usepackage{accents}%
\usepackage{mathtools}%
\usepackage{algorithm}
\usepackage{algpseudocode}

\setcitestyle{numbers}
\setcitestyle{square}
\usepackage{diagbox}
\usepackage[export]{adjustbox}
\usepackage{bm}
\usepackage[normalem]{ulem}

\journal{Knowledge-Based Systems}

\begin{document}
\begin{frontmatter}

\title{A Framework for Empowering Reinforcement Learning Agents with Causal Analysis: Enhancing Automated Cryptocurrency Trading}

\author[label1]{Rasoul Amirzadeh}
\ead{ramirzadeh@deakin.edu.au}

\author[label1]{Dhananjay Thiruvady\corref{cor1}}
\ead{dhananjay.thiruvady@deakin.edu.au}

\author[label1]{Asef Nazari}
\ead{asef.nazari@deakin.edu.au}

\author[label2]{Mong Shan Ee}
\ead{mong.e@deakin.edu.au}

\cortext[cor1]{Corresponding author.}
\address[label1]{School of  Information Technology, Deakin University, Geelong Waurn Ponds Campus, 3216, Australia}

\address[label2]{Deakin Business School, Deakin University, Burwood, Victoria, 3125, Australia}

\begin{abstract}
Despite advances in artificial intelligence-enhanced trading methods, developing a profitable automated trading system remains challenging in the rapidly evolving cryptocurrency market. This research focuses on developing a reinforcement learning (RL) framework to tackle the complexities of trading five prominent altcoins:
Binance Coin, Ethereum, Litecoin, Ripple, and Tether. To this end, we present the CausalReinforceNet~(CRN) framework, which integrates both Bayesian and dynamic Bayesian network techniques to empower the RL agent in trade decision-making. We develop two agents using the framework based on distinct RL algorithms to analyse performance compared to the Buy-and-Hold benchmark strategy and a baseline RL model. The results indicate that our framework surpasses both models in profitability, highlighting CRN's consistent superiority, although the level of effectiveness varies across different cryptocurrencies.

\end{abstract}


\end{frontmatter}

\section{Introduction}

Since the introduction of Bitcoin in Satoshi Nakamoto's paper~\citep{nakamoto2008peer}, the cryptocurrency market has grown substantially. From January 2019 to June 2024, the market capitalisation grew approximately 20 times, increasing from \$122.92 billion to \$2.6 trillion US dollars. Moreover, altcoins, which utilise blockchain protocols similar to Bitcoin but often with distinct features, now comprise a significant portion of the market~(around 45\% in June 2024).\footnote{www.tradingview.com} From around 1,700 notable altcoins in March 2017, the number has expanded to approximately 10,000 by early 2024, highlighting their increasing significance to investors.\footnote{www.statista.com} These facts underscore the importance of incorporating altcoins in cryptocurrency market research, even though Bitcoin remains an important element of the cryptocurrency ecosystem.



While empirical data highlights the growth and importance of cryptocurrencies, the market still encounters challenges similar to those of traditional financial assets, such as speculation and potential bubbles. Additionally, it faces unique challenges stemming from the attributes of the cryptocurrency market, such as decentralisation, utilisation of blockchain technology, 24/7 trading, and the need for regulatory acceptance—elements not typically supported by traditional financial systems. One significant challenge is the high volatility that offers profit opportunities but also has inherent risks of sudden downturns and potential losses. Therefore, providing investors with decision support system tools to leverage these opportunities while mitigating risks is crucial.



Financial markets are characterised by complexity, noise, and volatility~\citep{li2020stock}. Traditional statistical methods, such as the autoregressive integrated moving average~(ARIMA)~model, often fail to capture these dynamic characteristics. In contrast, artificial intelligence (AI) methods, particularly machine learning (ML) algorithms, have proven effective in designing trading systems~\citep{huang2019automated}, with applications such as algorithmic and high-frequency trading tools~\citep{sahu2023overview}. Among these applications, automated trading systems, which generate buy and sell signals based on predefined criteria, have become valuable tools in the cryptocurrency market, helping traders navigate the complex financial market~\citep{tran2023optimizing, puauna2019prediction}.\footnote{The terms `automated trading' and `algorithmic trading' are often used interchangeably in the literature~\citep{groth2014enable, gerig2017automated}.}



While ML algorithms, such as long short-term memory (LSTM)~\citep{swathi2022optimal} and artificial neural networks~\citep{liu2022quantum}, can effectively predict prices, trends, or volatility, they cannot construct fully automated trading systems~\citep{ma2021parallel}, and the lack of interpretability of these algorithms is a big issue. To effectively incorporate these algorithms, an extra logic layer is required to execute trading actions such as buying, holding, or selling based on their predictions~\citep{abdelkawy2021synchronous}.
Such an additional layer is demonstrated in the study by~\citet{omar2020stock}, where a module is created for stock market trading based on LSTM predictions.
Moreover, trading strategies that rely on predictions typically operate under fixed rules that remain unchanged once established. This inflexibility can pose inherent risks, as financial market trends are dynamic and frequently changing~\citep{chakole2021q, ansari2022deep}. However, reinforcement learning~(RL) offers several advantages over other ML algorithms. RL agents learn autonomously through interactions with the environment, without the need for pre-labelled data, unlike traditional supervised learning. They receive feedback in the form of rewards or penalties, allowing them to learn through trial and error~\citep{ikeda2022body,harewood2022q}. This capability enables the development of self-improving trading strategies, making RL a promising approach for building an automated trading system in the dynamic cryptocurrency market.

To address some challenges of automated trading systems within the cryptocurrency markets, this study proposes an RL-based trading framework. These challenges can be summarised as the following gaps extracted from the recent literature on using RL in such systems.
First, RL-based trading systems often use limited features, typically just the close price, as seen in studies~\cite{lucarelli2020deep}. Even when additional features like open price, high price, low price, and volume are used, proper feature engineering is not widely deployed in the literature. For instance, while some studies, such as \cite{weng2020portfolio} employ XGBoost with deep RL for portfolio trading of twenty cryptocurrencies, most studies, including \cite{kochliaridis2023combining, lee2018generating, schnaubelt2022deep}, rely on historical data and lack feature engineering. Therefore, in this research, we consider a wider range of features to improve the performance of the RL trading framework. In addition, we employ Bayesian networks~(BNs) for feature engineering to identify the most relevant features influencing cryptocurrency price movements, aiding the selection of appropriate features for the RL model. The causal nature of BNs, in addition, provides further interpretability to our proposed systems, helping the traders to have a better understanding of the most influential factors in price movements.

The second gap is the lack of integration with other ML algorithms to enhance the performance of RL-based systems in finance. While there have been recent efforts, such as a study by~\citet{cui2024integrating} that integrates deep autoencoder models with RL techniques to enhance financial risk forecasting in supply chain management, to the best of our knowledge, utilising other ML algorithms in conjunction with RL is neglected in the cryptocurrency domain. Thus, our proposed RL framework leverages price direction predictions using dynamic Bayesian networks (DBNs) to enhance the accuracy and effectiveness of financial trading decisions.

In summary, integrating other ML algorithms with RL in financial systems holds significant research value and practical importance. This study aims to enhance current RL trading models to address the challenges of high volatility in the cryptocurrency market by proposing a novel RL-based framework. The goal is to empower the RL model’s capability for effective trading in the cryptocurrency market. The main contributions of this paper are as follows.
\begin{itemize}
    \item Provides a broader understanding of the altcoin market by focusing on Binance Coin, Ethereum, Litecoin, Ripple, and Tether.
    \item Designs an RL-based framework by integrating two ML algorithms to improve trading performance.
    \item Incorporates BNs for feature engineering to identify the most relevant features influencing cryptocurrency price movements, improving the RL model’s feature selection.
    \item Utilises DBNs to enhance the RL agent’s ability to adjust trading decisions based on predicted price directions.
\end{itemize}

The remainder of this paper is structured as follows. 
Section~\ref{method} provides a brief introduction to RL. Our proposed RL automated trading system is outlined in Section~\ref{propsoed_system}, while Section~\ref{experimental} describes the implementation of our experiment and data analysis.
The results of the research are discussed in Section~\ref{Results}. Finally, Section~\ref{Conclusions} provides concluding remarks on the study and suggests potential directions for future research.

\section{An Overview of Reinforcement Learning \label{method}}
RL is a widely used branch of ML that is inspired by human decision-making processes, where learning is guided by the reward signals generated as a consequence of executed actions~\citep{littman2015reinforcement}.
RL aims to emulate the learning process of humans when they face an unprecedented task~\citep{lai2022towards}. To achieve this goal, RL agents are employed to interact with the environment by following a predefined procedure. In particular, the fundamental nature of RL is to learn through a process of trial and error. This means that the RL agent continually interacts with its environment, receiving feedback in the form of rewards and penalties, and learning to modify its behaviour accordingly. 

\begin{figure*}[h]
\centering
\includegraphics[width=0.8\linewidth,height=0.2\textheight,keepaspectratio]{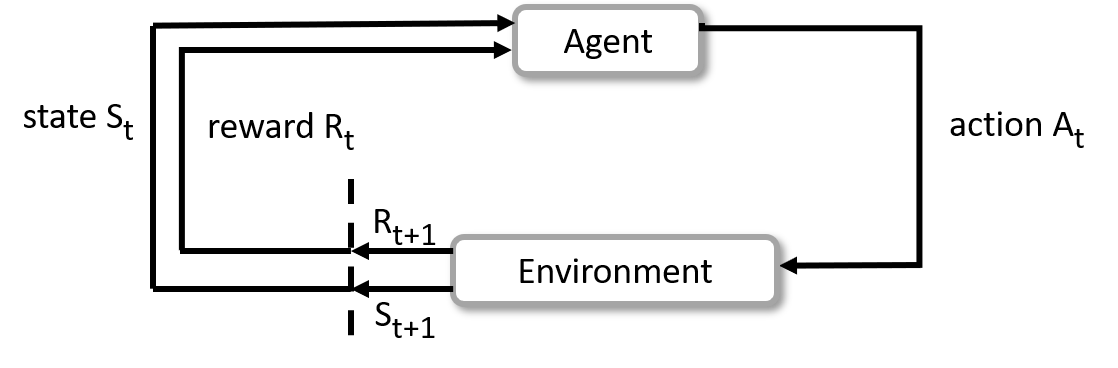}
\caption{\textbf{RL Classic Cycle}~--~The agent-environment interaction cycle is a continuous loop where an agent interacts with its environment in discrete time steps. At each time step, the agent observes the current state of the environment and selects an action based on its policy. After taking action, the agent receives a reward and uses it to optimise its policy. The ultimate goal for the agent is to develop a policy that allows it to maximise cumulative rewards over time.}  
\label{fig:MDP}
\end{figure*}

Figure~\ref{fig:MDP} shows the RL framework for agent–environment interactions. At each time step $t$, an agent finds itself in a particular state, represented as $S_t$, and selects an action, denoted as $A_t$. In response to this action, the agent receives a reward, denoted as $R_t$, and transitions to a new state, $S_{t+1}$. States, actions, and rewards are defined 
in the process of an RL agent's design.

An RL problem can be modelled as a five-tuple $(\mathcal{S}, \mathcal{A}, \uppi, \mathcal{R}, \lambda)$ Markov decision process (MDP). An RL algorithm is composed of the following components:

\begin{itemize}
\item []\textbf{States or observations $\mathcal{S}$}: A state serves as a complete representation of the conditions and circumstances within an environment. The agent obtains all relevant information from its states required to make decisions and interact effectively with the environment. 
\item []\textbf{Action spaces $\mathcal{A}$}: In every environment, a range of actions can be performed. This set of acceptable actions within a specific environment is referred to as the action space. The nature of the action space can differ based on whether it is discrete or continuous. 
As opposed to a continuous action space, an agent has a finite number of available actions in a discrete action space.
\item []\textbf{Policy $\uppi$}: A policy $\pi$ is the way an agent behaves at a given time. It determines the action that has to be decided by the agent when the agent is in a particular state within the environment. In a probabilistic context, a policy maps the current state of the environment to a set of probabilities associated with each action in the action space. These probabilities indicate the likelihood or chance of the agent performing each possible action. 
\item []  \textbf{Rewards $\mathcal{R}$}:  The reward is a unique component of RL that represents the immediate gain or benefits an agent receives for selecting a specific action in the current state, resulting in a transition to the following state. 
 \item []  \textbf{Discount factor $\lambda$}: The discount factor is a parameter that determines the relative importance of future rewards compared to immediate rewards. The discount factor influences the agent's decision-making by assigning weight to future rewards. A value of 0 indicates that the agent only considers immediate rewards and disregards any potential long-term consequences. In contrast, $\lambda = 1$ means that the agent evaluates its actions based on cumulative rewards in the future.

\end{itemize}

RL algorithms enable systems to learn optimal actions through interaction with their environment. There are generally two types of RL algorithms: model-based and model-free. Model-based algorithms have the knowledge of the transition matrix and reward function, allowing the agent to plan ahead and make decisions based on a range of possible options. On the other hand, model-free methods lack knowledge of the transition matrix and reward function, relying on previous experience to estimate state-action pairs~\citep{cho2022reinforcement}. Model-based algorithms in RL are computationally expensive as they require learning an accurate environment model, which can be a challenging task in real-world applications. Hence, model-free algorithms are more popular since they are less computationally expensive~\citep{touzani2021controlling}.

Model-free algorithms are classified into value-based and policy-based. Value-based methods aim to estimate a {\it value function}, which assigns a value to each state or state-action pair. The value function represents the expected return 
that can be obtained from a particular state or state-action pair~\citep{zhang2018dynamic}. 
On the other hand, policy-based methods directly learn the policy, which is a mapping from states to actions~\citep{liu2021balancing}. 
Instead of estimating the value function, policy-based algorithms focus on learning the policy directly~\citep{dang2022towards}.

\section{A Comprehensive Guide to the CausalReinforceNet Framework}
\label{propsoed_system}

\begin{figure*}[h]
  \centering
\includegraphics[ width=0.99\linewidth,height=0.9\textheight,keepaspectratio]{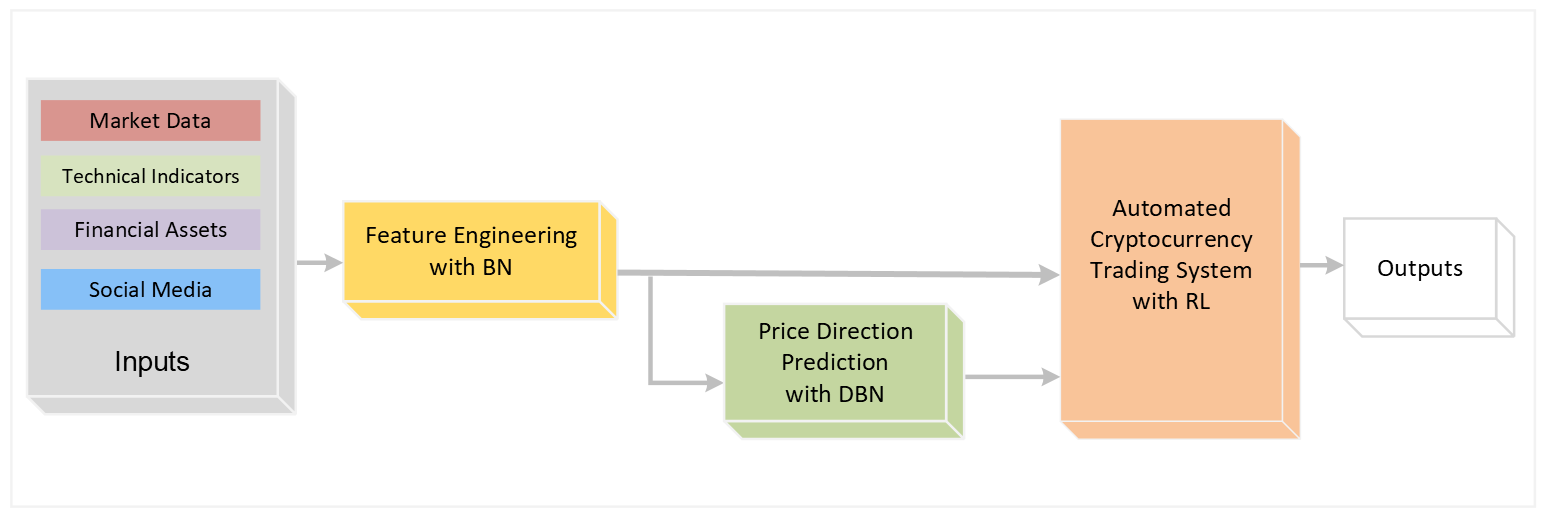}
\caption{\textbf{CRN Architecture}~--~The framework comprises four essential elements. The first category encompasses input features containing market data, technical indicators, financial assets, and social media. The second element consists of a BN module, which is responsible for the feature engineering of the inputs. The third element is the DBN, which generates daily price directions based on the selected features. Finally, the fourth element represents the automated RL-based trading system.}
  \label{fig:Rl_model}
\end{figure*}

This section outlines the development process of the CRN  framework for automated trading. To optimise RL performance, careful design of its components such as state representation, action space, reward function, and algorithm selection is crucial. CRN serves as the architectural foundation for our RL-based trading system. As depicted in Figure~\ref{fig:Rl_model}, CRN comprises four key components. This figure illustrates the system's input features and how BN, DBN, and RL methods interact to enhance the agent's decision-making capabilities.


The system's initial component (from left to right in Figure~\ref{fig:Rl_model}) involves input data, which is categorised into four distinct feature sets. Subsequently, the BN model selects the most influential features from these sets for each individual altcoin. The DBN component then utilises these selected features to generate price predictions for each cryptocurrency. Finally, the RL component integrates insights from both the BN and DBN to execute automated trading decisions.


Different elements of the RL component of our CRN are outlined as follows.
\subsection{States set $\mathcal{S}$} 
In an RL-based trading system, observations represent the current state of the market environment perceived by the RL agent. These observations guide the decision-making process and actions of the agent within the trading system. However, developing an effective policy depends on extracting informative states from high-dimensional observations~\citep{killian2020empirical}, which is particularly challenging in finance, where numerous factors, often characterised by noise and complexity, contribute to these observations.
When applied to real-world tasks, it is necessary to construct a suitable state space to mitigate the dimensionality problem~\citep{hashemzadeh2018exploiting, arai2018reinforcement}. Therefore, employing feature selection or dimensionality reduction techniques can improve RL algorithms' efficiency and computational effectiveness~\citep{weng2020portfolio}.

In this research, we utilise various features that can aid in predicting prices in financial markets. These features are derived from sources of information available both within and outside the cryptocurrency market. The selected group of features is utilised as the state representation within the trading system for each altcoin.
The first group comprises OHLCV data, encompassing basic price information 
of the altcoins. OHLCV presents the five key data points over a specified period. Notably, the closing price is often regarded as the most significant metric by many financial market participants.
The second group is the combination of OHLCV with ten technical indicators, including AD (Accumulation/Distribution), BBands (Bollinger Bands), EMA (Exponential Moving Average), OBV (On-Balance Volume), MACD (Moving Average Convergence/Divergence), NATR (Normalised Average True Range), RSI (Relative Strength Indicator), SMA (Simple Moving Average), and Stoch (Stochastic Oscillator). These indicators provide insights into various aspects of the financial market, such as trend strength, the potential reversal of price trends, and price volatility \citep{alonso2020convolution, srivastava2021deep}.

The third group of features combines OHLCV data with five traditional financial assets, including gold, the US dollar index (USDX), Standard \& Poor's 500 Index (S\&P 500), MSCI, and West Texas Intermediate (WTI), and the number of daily tweets associated with each altcoin. 
While these features are not directly linked to the cryptocurrency market, they can influence cryptocurrency prices. The fourth group integrates all the aforementioned features into a single group.
The combination of these features with OHLCV enables our RL agent to consider broader market influences and the impact of social media on our analysis. A brief overview of these features can be found in Table~\ref{tab:features}.

\begingroup
\renewcommand{\arraystretch}{1.2} 
\begin{table}[]
\centering
\caption{\textbf{Feature description}~-~The table provides a short description of selected features, types, and the time window.}
\label{tab:features}
\begin{tabular*}{\textwidth}{p{2.2pc}p{3.9pc}p{13pc}p{9.7pc}}
\hline
\multicolumn{1}{l}{\textbf{Type}} & \textbf{Feature} & \multicolumn{1}{c}{\textbf{Description}} & \multicolumn{1}{l}{\textbf{Time window}} \\ 
\hline
\noalign{\smallskip}
\multicolumn{1}{l}{\multirow{5}{*}{OHLVC}} & Close price & Last traded price in an interval & Daily \\
& Low price & Lowest trading price in an interval & Daily \\
& High price & Highest trading price in an interval & Daily \\
& Open price & First traded price in an interval & Daily \\
& Volume & The total number of cryptocurrencies traded  in a trading interval& Daily \\ 
\hline
\noalign{\smallskip}
\multicolumn{1}{l}{\multirow{5}{*}{\shortstack{Macro\\financial}}}  & Gold & The gold spot market price in USD & Daily \\
& MSCI & Market capitalisation-weighted index of 1,546 global companies & Daily \\
& S\&P500 & Market capitalisation-weighted index of top 500 US companies & Daily \\
& USDX & US dollar value relative to six foreign currencies & Daily \\
& WTI & A prevalent oil price benchmark & Daily \\ 
\hline
\noalign{\smallskip}
\multicolumn{1}{l}{\multirow{2}{*}{\shortstack{Social\\media}}}  & Tweet No. & The number of daily tweets associated with a cryptocurrency  & Daily \\ 
\hline
\noalign{\smallskip}
\multirow{8}{*}{\shortstack{Technical\\indicators}} & AD & Uses volume and price for money flow & Last Period \\
& BBands & Three lines, central SMA with 2 standard deviations boundaries & 5 days \\
& EMA & Weighted moving average to reduce the data lag & 10 days \\
& MACD & Measures two EMAs & Fast period=12,\newline Slow period=26,\newline Signal period=9 \\
& NATR & Metric measuring volatility & 14 day \\
& OBV & Cumulative indicator of buying and selling pressure & Last period \\
& RSI & Measures overbought and oversold conditions & 14 days \\
& SMA & Simple moving average of data points & 10 days \\
& Stoch & Range-bound momentum indicator to determine overbought and oversold situations & 14 days \\  
\hline
\end{tabular*}
\end{table}
\endgroup

 We utilise BNs for feature engineering where they are used to systematically explore different combinations of features to identify those features which are most influential in predicting altcoin prices. This process, termed `Feature Engineering with BN',  
 is depicted in Figure~\ref{fig:Rl_model}. The process involves determining the impact of these feature groups on altcoin price movements.  BNs evaluate the predictive power of each feature group and identify the optimal combination of features that achieve high prediction performance for each cryptocurrency.
Once these influential features are identified through iterative testing and validation within the BN component of the framework, we select the optimal combination of features (outlined in Table~\ref{tab:obs}).


In addition to the observations in Table~\ref{tab:obs}  and features utilised in our framework, we enhance the observations of the RL agent by incorporating a specific type of market observation which is predictions for daily closing price directions of altcoins from a DBN.
The framework leverages the DBN predictions to provide a distinct layer of information that enriches the agent’s decision-making process.
In particular, the DBN generates probabilistic predictions of daily prices of altcoins, meaning that for each day, it presents the probabilities of a price increase (Up) and a decrease (Down). It utilises the last five days of data to provide probabilistic predictions for Up and Down movements
, which are then fed into the RL agent. This enhancement enriches the RL agent's observations and ensures a more informed approach to decision-making in our model. This process is visually represented by `Price Direction Prediction with DBN' in Figure~\ref{fig:Rl_model}. 

It should be noted that each altcoin is influenced by distinct factors affecting its price movements, as shown in Table~\ref{tab:obs}, and accordingly, DBNs use these related features for each altcoin to predict the daily price directions. Particularly, while daily closing price predictions for Binance Coin and Ripple are produced using both OHLCV and technical indicators, the DBN generates the predictions for Ethereum and Tether using only OHLCV. Notably, the DBN predicts the daily closing price of Litecoin based on a combination of OHLCV, tweet numbers, and financial factors.
These DBN predictions are integrated into the observations of all altcoins to enhance our agent's trading strategy, offering valuable insights into market trends and potential trading opportunities.\footnote{A more detailed explanation regarding BN and the DBN processes can be found in~\cite{amirzadeh2023modelling, amirzadeh2023causal}.}
\begin{table*}[h]
\caption{\textbf{Agent Observations}~--~Observations selected for each altcoin with feature engineering using BN.}
\vspace{-0.2cm}
\label{tab:obs}
\resizebox{0.99\columnwidth}{!}{
\renewcommand{\arraystretch}{1.3}
\begin{tabular}{@{}llc@{}}
\toprule
\textbf{Cryptocurrency} & \multicolumn{1}{c}{\textbf{Observations}} & \textbf{Number of Features} \\ \midrule

Binance Coin & OHLCV and technical indicators& 15 \\ 
Ethereum     & OHLCV                                      & 5  \\ 
Litecoin     & OHLCV, financial factors, and tweet number                                     & 11  \\ Ripple       & OHLCV  and technical indicators & 15 \\ 
Tether     &  OHLCV & 5 \\ \bottomrule
\end{tabular}
}
\end{table*}

\subsection{ Action $\mathcal{A}$}

In a trading procedure, at least two principal decisions are evaluated at each timestep. The first decision pertains to the trading action, which is influenced by current market data: a trader may decide to buy if the price is anticipated to ascend, sell if a decline is expected, or hold in cases of market uncertainty. The second decision concerns the allocation of capital to each trade, commonly referred to as the position size. This involves determining the number of units of a financial asset that a trader or investor buys or sells in a single transaction. The position size can be in various forms, such as 
a set number of shares in stock trading, or a specific money amount. This parameter is affected by various strategic factors, including the trader's risk tolerance and portfolio management objectives.

In our framework, the agent's action space reflects these two important trading decisions. The agent can select from actions including buy, hold, or sell, based on the perceived market conditions. Additionally, it can decide on the position size for each trade. Unlike typical RL trading approaches (e.g.~\cite{wu2020adaptive}) that trade a nominal quantity of one share per transaction, our framework allows the agent to dynamically adjust the number of assets being bought or sold. By incorporating this additional layer of decision-making, our agent can optimise its trading performance and adapt to different market conditions.

Furthermore, in designing a trading system, various rules can be established to serve different purposes, such as mitigating risk. For instance, some research defines a maximum number of open positions for portfolio management, preventing the agent from opening new positions once this limit is reached~\cite{maleki2023risk}. Given the high volatility in the cryptocurrency market, we design our agent to adopt a more conservative strategy. This strategy employs a hybrid approach that combines rule-based strategies with RL capabilities.
We set specific rules regarding the amount of cryptocurrency that can be sold and the amount of money that can be allocated for purchasing. This approach helps manage risk more effectively by combining the adaptiveness of RL with the stability of predefined trading limits. In particular, the trading strategy is constrained to invest within a certain range. Through experimentation, we discovered that setting thresholds between 40\% and 60\% of the available balance yields higher returns.
The purchased cryptocurrency is then stored in the agent's wallet. Limiting trades to a maximum of 60\% of available balance minimises risk by preventing overexposure to market volatility from any single trade, while a 40\% minimum threshold allows having enough money for future trading opportunities. Moreover, during a sell action, the agent can sell between 40\% (lower threshold) and 60\% (upper threshold) of the cryptocurrency available in its wallet.
This strategy ensures that a significant portion of the hold cryptocurrencies is preserved, mitigating the impact of sudden market downturns. 
Although implementing these restrictions on the buying and selling actions may potentially decrease the profitability of our agent, we aim to mitigate the risks associated with large and sudden market fluctuations while still allowing the agent to participate actively in trading activities. 
It is important to mention that individual traders may consider different thresholds based on various factors, such as risk management strategies and risk appetite.

Furthermore, to protect our agent from sudden movements in the cryptocurrency market, we use the price direction predictions from the DBN to guide its trading decisions. In particular, when the DBN indicates a Down price direction with a probability exceeding 80\% (indicating a strong downward market condition ), the agent is programmed to sell 75\% the maximised threshold) of the cryptocurrency in its wallet. Conversely, if the DBN signals an Up price direction with an 80\% probability (suggesting a strong upward market condition), the agent will allocate 75\% of its available balance as the trade position size to purchase more cryptocurrency. These adaptive rules enable the agent to adjust its position sizes based on the market conditions indicated by the DBN predictions. 

Finally, we consider the transaction cost associated with each trade, which directly impacts the overall profitability of the agent.  Therefore, we incorporate a transaction cost of 0.1\% based on the position size of every trade, which is deducted from the balance managed by the agent to ensure a realistic evaluation of the trading strategy's performance.
Transaction fees can vary in different exchanges, while 0.1\% is the most commonly occurring fee. A list of top transaction fees in most known cryptocurrency exchanges can be found in~\cite{amirzadeh2022applying}.

The pseudo-code outlined in Algorithm~\ref{alg:position-sizing} shows the step-by-step actions executed by our trading agent within each time step, as discussed earlier. 
The position-sizing algorithm outlines the agent's strategy to determine the size of trades when buying or selling. Specifically, for a sell action, the algorithm sells 75\% of the available wallet amount (WA) if there is an 80\% or higher likelihood of a downward price movement. Otherwise, it sells a position size ranging from 40\% to 60\% of WA. In contrast, for a buy action, the agent invests 75\% of the available balance (B) if there is at least an 80\% probability of an upward price movement. Otherwise, the buying position size is between 40\% and 60\% of B. Transaction costs (TC) are then calculated based on the selected trading position. The algorithm performs a hold action if neither buy nor sell actions are chosen.

\begin{algorithm}
\caption{Trading Position Sizing Algorithm}
\label{alg:position-sizing}

\textbf{Inputs:}
\begin{itemize}
  \item \textbf{WA}: Amount of available coins in the wallet
  \item \textbf{B}: Available balance for trading
  \item \textbf{TC}: Transaction cost per trade
\end{itemize}

\begin{algorithmic}[1]
  \If {$\text{Action = Sell}$}
    \If{$\text{Probability of `Down' price direction by DBN} \geq 80\%$}
      \State Sell Position Size = 75\%
    \Else
      \State Sell Position Size = Between 40\% and 60\% of WA
    \EndIf
    \State TC = WA $\times$ Sell Position Size $\times$ 0.1
    \State WA = WA $\times$ (1 - Sell Position Size)
     \State B = B - TC
  \ElsIf {$\text{Action = Buy}$}
    \If{$\text{Probability of `Up' price direction by DBN} \geq 80\%$}
      \State Buy Position Size = 75\%
    \Else
      \State Buy Position Size = Between 40\% and 60\% of B
    \EndIf
    \State TC = B $\times$ Buy Position Size $\times$ 0.1
\State B = B - (B $\times$ Buy Position Size + TC)
  \Else
    \State \Return {Hold}
  \EndIf
\end{algorithmic}
\end{algorithm}

\subsection{ Reward~$\mathcal{R}$}

The primary goal of a trading RL agent is to maximise its profitability. While many studies utilise the return on investment (ROI)\footnote{The return on investment measures the profitability of an investment by comparing the return to the original cost.} as the sole criterion for the reward function in RL trading systems, this approach often overlooks crucial elements such as market stability and risk factors \citep{wu2021portfolio}.\footnote{Financial market stability, although a multifaceted topic, can be defined as a state wherein a financial system is flexible and adaptable for effective risk management and absorbing shocks~\citep{houben2004toward}.}

To address these limitations, we incorporate the Sharpe Ratio~(SR) alongside ROI into our reward function.
SR quantifies the expected return per unit of risk undertaken. It provides a measure of the risk-adjusted performance of an investment or trading strategy, and is defined as follows: 
$$ {S_p} = \frac{\varrho_p - \varrho_f}{\sigma_p}$$

where $\varrho_p$ is average return, $\varrho_f$ is the risk-free rate, and $\sigma_p$ is standard deviation of investment return. In contrast to studies that assume a zero risk-free rate, we incorporate a realistic rate of 3.4\%\footnote{A government bond is a security issued by a government to support its spending. These bonds generally offer periodic interest payments and are considered low-risk due to the backing of the government. The 3.4\% rate refers to the interest rate of the Australian 10-year Government Bond in 2023.} enhancing the accuracy of our model in reflecting real-world conditions.

Our reward structure  combines weighted ROI and SR metrics to simultaneously promote profitability and enhance risk management. We adapt a method proposed by~\citet{lucarelli2019deep}, which suggests  balancing these factors to optimise trading strategies.
$R_{SR}$ function at time $t$ is defined as:

\[
R_{SR} =
\begin{cases}
    10 & \text{if } S_{p_t} \geq 4 \\
    4 & \text{if } 1 \leq S_{p_t} \leq 4 \\
    1 & \text{if } 0 \leq S_{p_t} \leq 1 \\
    0 & \text{if } S_{p_t} = 0 \\
   -1 & \text{if } -1 \leq S_{p_t} \leq 0 \\
   -4 & \text{if } -4 \leq S_{p_t} \leq -1 \\
  -10 & \text{if } S_{p_t} \leq -4 \\
\end{cases}
\]

The~$R_{ROI}$ function is defined as:

\[
R_{ROI} =
\begin{cases}
10 & \text{if } ROI \geq 0.5 \\
4 & \text{if } 0.2 \leq ROI \leq 0.5 \\
1 & \text{if } 0.1 \leq ROI \leq 0.2 \\
0 & \text{if } ROI = 0 \\
-4 & \text{if } -0.2 \leq ROI \leq 0 \\
-10 & \text{if } ROI \leq -0.2 \\
\end{cases}
\]

The use of a non-linear step function in the reward calculation is driven by the need to enhance the RL agent's sensitivity to shifts in risk-adjusted returns and to penalise undesirable trading actions more significantly. By implementing threshold-based steps, we can establish clear incentives for the agent to seek higher risk-adjusted returns while avoiding overly risky strategies.

Experimentation with different weight distributions revealed that allocating 70\% to $R_{ROI}$ 
and 30\% allocation to $R_{SR}$ yields a higher overall return. Therefore, the final reward function is defined as below: 
$$\text{$\mathcal{R}$} = 0.7 * \text{$R_{ROI}$} + 0.3 * \text{$R_{SR}$}$$

\subsection{RL algorithm}

There are various RL algorithms available for training RL agents. However, as there is no universal solution, it remains challenging to identify which algorithm is best suited for a particular task~\citep{di2022deep}. Our study examines two commonly used RL algorithms: Proximal Policy Optimisation and Deep Deterministic Policy Gradient~(DDPG).
These two algorithms are both model-free; however, they possess distinct key differences in their approaches and applications.

Proximal Policy Optimisation~\citep{schulman2017proximal} is an on-policy algorithm that aims to optimise a policy by taking the biggest possible improvement step on a policy while using the current experience without moving so far from the current policy~\citep{melo2021learning}. PPO, developed by OpenAI, supports a broad range of state and action representations~\cite {lee2022reward}. Moreover, this algorithm, adopted as the default algorithm by OpenAI, is used to train the RL agent due to its simplicity, generality, and low sample complexity. Unlike most ML algorithms requiring extensive hyperparameter tuning, PPO can create effective results with default settings or minimal adjustments~\citep{kegenbekov2021adaptive}.

As the second algorithm under investigation, we examine Deep Deterministic Policy Gradient~(DDPG)~\citep{silver2014deterministic}. It is a deep RL method that combines RL with deep learning strategies to enhance the performance of agents in complex environments. DDPG approach employs deep neural networks, which comprise multiple hidden layers populated with numerous units each, to facilitate learning~\citep{wang2023deep}. This is an off-policy algorithm with better real-time adaptability and can only work in environments with continuous action spaces~\citep{li2021efficient}.

These algorithms utilise a continuous state and action space, which makes them particularly suitable for the trading rules within our system. Moreover, they demonstrate great potential in this complex environment of the financial markets~\citep{conegundes2020beating, li2021enhancing}. Therefore, we develop two distinct agents of the CRN framework by modifying the RL component based on these algorithms. One version utilises the PPO algorithm and is denoted as the {$CRN_{PPO}$} agent, while the other employs the DDPG algorithm, termed the {$CRN_{DDPG}$} agent. Through analysis of these algorithms within the context of our proposed framework, we aim to determine the more suitable approach, considering the unique features of each algorithm.

\section{Experimental Design}\label{experimental}
To conduct the experiments, the data was divided into training and testing sets using a 67\% to 33\% ratio, which is a commonly used split ratio~\citep{lyons2018comparison}. 
In order to assess the effectiveness of our proposed trading model, we compare its performance to a common benchmark trading strategy used by many investors called Buy-and-Hold. This traditional investment strategy involves purchasing and holding assets for an extended period without actively trading. We conduct an additional experiment to investigate the influence of BNs and DBNs within our framework. Specifically, we develop a baseline RL model for each algorithm that only utilises OHLCV features without incorporating any causal feature engineering. We select the OHLCV group as the base feature for the RL model because it is commonly used across all feature groups. By isolating the OHLCV features, our objective is to understand how modifications to the framework impact the overall system performance. Moreover, the ROI metric is used to evaluate the performance of our agent and Buy-and-Hold strategy. ROI measures the profitability of an investment relative to the initial investment amount. In our examination, we calculate the ROI of our model by selling all available cryptocurrencies held in the agent's wallet at the final time step.
Basic ROI calculations do not account for the duration of the investment. Considering the duration of the investment is essential because it provides valuable insights into the consistency and sustainability of trading strategies. Therefore, we also calculate the annual ROI, which provides an average yearly return on the investment. By considering both ROI and annual ROI, we gain a more comprehensive understanding of both the overall profitability and the year-over-year returns generated by implemented trading strategies.

We conduct experiments on a Windows 10 Enterprise system with an Intel i7-Core CPU and 16 GB of RAM. Our study is implemented in Python, using the Stable Baselines3 library for RL algorithms and the pandas-ts library for computing technical indicators. In this study, we use the default hyperparameter settings for DRL models. This is because the aim is to primarily assess the framework's performance across various cryptocurrencies without introducing the complexities associated with customising hyperparameters for each individual altcoin.
All abbreviations used in the study are provided in Table~\ref{tab:abbrv} to facilitate ease of reference.

\begin{table}[h]
\centering
\caption{\textbf{Abbreviations}~--~The table provides an alphabetical list of abbreviations and their corresponding explanations, which are frequently used in the study.}
\label{tab:abbrv} \resizebox{0.99\columnwidth}{!}{%
 \renewcommand{\arraystretch}{1.5}
\begin{tabular}{|l|l|l|l|}
\hline
\multicolumn{1}{|c|}{\textbf{Abbreviation}} & \multicolumn{1}{c|}{\textbf{Explanation}} & \textbf{Abbreviation} & \multicolumn{1}{c|}{\textbf{Explanation}} \\ \hline
AI & Artificial Intelligence & ML & Machine Learning \\ \hline
BNs & Bayesian Network & PPO & Proximal Policy Optimisation \\ \hline
CRN & CausalReinforceNet & RL & Reinforcement Learning \\ \hline
$CRN_{DDPG}$ & CRN implemented by DDPG algorithm & ROI & Return on Investment \\ \hline
$CRN_{PPO}$ & CRN implemented by PPO algorithm & $R_{ROI}$ & ROI Reward Function \\ \hline
DBNs & Dynamic Bayesian Network & $R_{SR}$ & SR Reward Function \\ \hline
DDPG & Deep Deterministic Policy Gradient & SR & Sharpe Ratio \\ \hline
\end{tabular}
}
\end{table}

\subsection{Data}
The research investigates five cryptocurrencies: Binance Coin, Ethereum, Litecoin, Ripple, and Tether. These altcoins have consistently been among the top 10 cryptocurrencies in terms of market capitalisation, highlighting their popularity and influence in the cryptocurrency market. Their market capitalisation represents a substantial portion of the cryptocurrency market, over 30\% of the overall market in the middle of  2023.
Moreover, we include altcoins with a minimum of 1,900 daily data records, equivalent to more than six years, to enhance our analysis's robustness and statistical significance.

The selection of a time interval, as the characteristics of temporal data, can influence analytical outcomes~\citep{adhikari2012mining}. Trading rules also vary based on the timeframe preferences of financial market participants, which are tailored to suit specific predictive applications. We select daily data for our study primarily because this approach not only helps avoid computational challenges but also reduces the excessive market noise inherent in high-frequency data analysis. Moreover, processing tweet data at high frequency is a complex task, and such data is not available from our source, including bitinfocharts.com. 

The daily price data for both altcoins and traditional financial assets are collected between January 2018 and April 2023 from Yahoo Finance\footnote{https://finance.yahoo.com/} using the yfinance Python package. Furthermore, the daily tweet number for the related altcoins is extracted from bitinfocharts.com.
Table~\ref{tab:df_description} presents the descriptive statistics of the close price data for the studied altcoins.

\begin{table*}[ht]
 \centering
 \caption{\textbf{Summary of Altcoin Price}~--~ The table presents summary statistics for the dataset of daily close prices of cryptocurrencies between January 2018 and April 2023
  with a total observations of 1945 records
 are presented in the table. The columns provide information about statistical measures such as mean, standard deviation, minimum, maximum, and median.}
 \vspace{-0.1cm}
 \label{tab:df_description}
 \resizebox{\columnwidth}{!}{%
 \renewcommand{\arraystretch}{}
 \begin{tabular}{lrrrrr}
 \cmidrule{2-6}
  \textbf{} & \textbf{Mean} & \textbf{Std. Dev.} & \textbf{Min.} &\textbf{Median}&\textbf{Max.} \\  \cmidrule{2-6}
    \textbf{Binance coin} & 160.61 &  182.14 & 4.53& 29.02 &675.68 \\ 
  \textbf{Ethereum} & 1179.13 &  1170.65 & 84.31 &609.82&4812.09\\   
  \textbf{Litecoin}  & 98.89 &  59.36 & 23.46 &77.52 &386.45\\  
  \textbf{Ripple}  & 0.52 &  0.35 & 0.14 &0.39 &3.38\\ 
  \textbf{Tether}  & 1.00 &  0.01 & 0.97 &1.00 &1.05\\ \bottomrule
\end{tabular}
}
\end{table*}

\begin{figure*}[h]
  \centering
\includegraphics[width=0.85\linewidth,height=0.9\textheight,keepaspectratio]{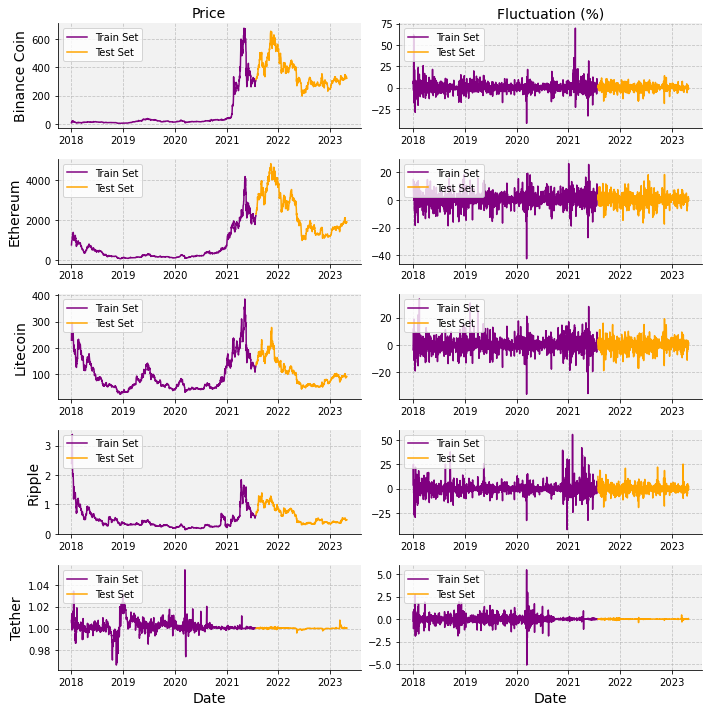}
\caption{\textbf{Price Plots}~--~The plots display the price patterns and fluctuations of studied cryptocurrencies between January 2018 and April 2023. }
  \label{fig:coins_plots}
\end{figure*}

Figure~\ref{fig:coins_plots} provides a visual representation of the price and fluctuation patterns of the cryptocurrencies of the research. This visualisation facilitates the analysis of price trends and volatility for each cryptocurrency over time. Moreover, the training and test sets are visually differentiated through distinct colours, which helps to understand their different characteristics and dynamics.

There are several notable observations from the plots. All coins exhibit both positive and negative fluctuations in their price movements. Binance Coin, Ethereum, Ripple, and Litecoin demonstrate significant price fluctuations, ranging approximately between -25\% and 75\%. On the other hand, Tether, being a stablecoin designed to maintain a stable value pegged to the US dollar, does not exhibit significant price fluctuations compared to other cryptocurrencies. These observations highlight the dynamic nature of the cryptocurrency market and the varying price behaviours demonstrated by different altcoins. 
Additionally, during both the training and test periods, there are extreme downward and upward market conditions that present a challenging situation for our agent.

Furthermore, the period of 2021 experienced a considerably higher level of variation compared to other periods covered in the study. For instance, Ethereum showed high variances of 53,225.206 and 905,922.229 before and after 2021, respectively. This considerable variance increase indicates the significant fluctuation and volatility associated with Ethereum during the year 2021. As a result, high prediction error is expected due to the growing variances in the data.

\section{Results and Discussions}\label{Results}
In this section, we present the results of our experiment.  As suggested by~\citet{kong2023empirical}, a single experiment may not provide conclusive evidence of a trading system performance; therefore, we conduct five experiments. Then, the average value for various metrics is calculated for altcoins. In this section, we initially analyse the two RL agents by comparing them with the Buy-and-Hold strategy.
Subsequently, to delve deeper into the effectiveness of the CRN framework, we expand the analysis to compare the two agents with RL-based models. When analysing the results, one key consideration is that the test period overlaps with the COVID-19 pandemic and the subsequent post-COVID phase. Throughout the COVID-19 period, many financial and economic factors experienced extreme changes, resulting in noticeable instability across various aspects of the economy, including inflation, high interest rates, and supply chain disruptions. Although these volatile conditions can inevitably influence the performance of the trading system, it is still important to investigate the robustness of our trading system during this period. It ensures that our system can not only survive but thrive in volatile financial conditions, eventually making it practical to be deployed by real-world traders. Additionally, the cryptocurrency market lacks extensive historical data, which makes excluding this turbulent period from our analysis impractical.

\begin{figure*}[h]
  \centering
\includegraphics[width=0.95\linewidth, height=0.8\textheight, keepaspectratio]{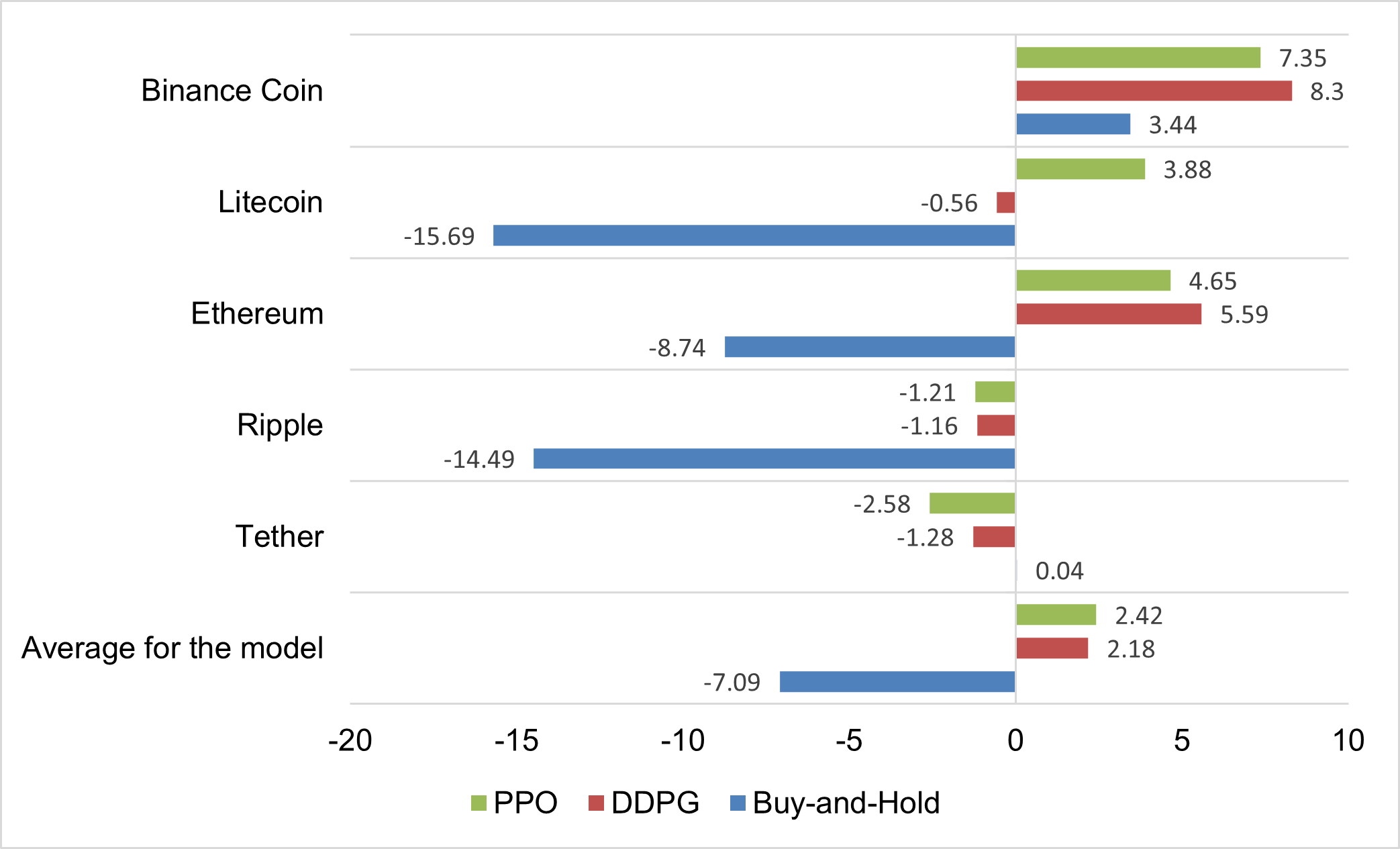}
\caption{\textbf{Performance Comparison Visualisation}~--~The chart compares the annual ROI of studied cryptocurrencies as detailed in Table~\ref{tab:preds}.}
  \label{fig:performance}
\end{figure*}

The results are illustrated in Table~\ref{tab:preds}, which outlines ROI and the annualised ROI of our trading system for three different strategies: two RL agents of {$CRN_{PPO}$} and {$CRN_{DDPG}$} in addition to the Buy-and-Hold approach. To facilitate the comparison, the performance of models based on annual ROI is visually presented in Figure~\ref{fig:performance}.

\begin{table*}[ht]
\caption{An evaluation of the three approaches on five altcoins based on the percentage of ROI and annual ROI for PPO and DDPG agents. Numbers in parentheses show the standard deviation. `Ann.' stands for `Annual.' }
 \vspace{-0.1cm}
\label{tab:preds}
  \resizebox{0.99\columnwidth}{!}{%
 \renewcommand{\arraystretch}{1.25}
\begin{tabular}{lrrrrrr}
\cmidrule{2-7}
\textbf{} & \multicolumn{2}{c}{\textbf{CRN$_{\textbf{PPO}}$}} & \multicolumn{2}{c}{\textbf{CRN$_{\textbf{DDPG}}$}} & \multicolumn{2}{c}{\textbf{Buy-and-Hold}} \\ \cmidrule{2-7}
 &\multicolumn{1}{c}{ROI} & \multicolumn{1}{c}{Ann.ROI} & \multicolumn{1}{c}{ROI} & \multicolumn{1}{c}{Ann.ROI} & ROI & \multicolumn{1}{c}{Ann.ROI} \\ \hline
\textbf{Binance Coin} & 12.93~(30.34)  & 7.35~(17.25)  & 14.59~(17.18) & 8.30~(9.77) & 5.92 & 3.44 \\ \hline
\textbf{Ethereum} & 8.19~(44.56) & 4.65~(25.34)  & 10.46~(40.15) & 5.59~(22.83) & -14.40 & -8.74 \\ \hline
\textbf{Litecoin} & 5.13~(26.06)  &  3.88~(19.70) & -0.74~(13.43) & -0.56~(10.15) & -25.18 & -15.69 \\ \hline
\textbf{Ripple} & -2.13~(15.80) & -1.21~(8.99) & -2.04~(36.24) &-1.16~(20.61)  & -23.36 & -14.49  \\ \hline
\textbf{Tether} & -4.54~(0.18)  & -2.58~(0.10)  & -2.26~(1.88)  &-1.28~(1.07)  & 0.06 & 0.04 \\ \hline
Average & 3.92 & 2.42 & 4.00 & 2.18& -11.39 & -7.09 \\ \bottomrule
\end{tabular}
}
\end{table*}

Considering our conservative trading strategy designed
to mitigate the inherent risks of the cryptocurrency market
, our agent generally performs well in generating positive results, particularly in challenging situations during the test period.
Both RL agents
mostly outperforms the Buy-and-Hold strategy. For instance,
{$CRN_{PPO}$} agent
generates a higher ROI and annual ROI of 12.93\% and 7.35\%, respectively, for Binance Coin, compared to the Buy-and-Hold. 
Similarly, while the Buy-and-Hold strategy results in negative ROI for Ethereum, this RL agent manages to gain a positive ROI and annual ROI of 8.19\% and 4.65\%. Litecoin also produces positive ROI and annual ROI results of 5.13\% and 3.88\%. However, {{$CRN_{PPO}$ } performs poorly with Ripple and Tether, generating negative ROI. Despite the negative returns, the {$CRN_{PPO}$ agent's losses are considerably lower than the Buy-and-Hold strategy for Ripple.
On the other hand, the $CRN_{DDPG}$ agent outperforms the Buy-and-Hold strategy. It generates higher ROIs for both Binance Coin and Ethereum, 14.59\% and 10.46\%, respectively. However, it does not perform well on Litecoin, Ripple, and Tether. Even with negative returns, the loss of investment of $CRN_{DDPG}$  is still substantially lower than the Buy-and-Hold strategy.

From the information in Table~\ref{tab:preds}, we can gain valuable insights into the performance comparison of PPO and DDPG RL algorithms.
The RL agent implemented with DDPG algorithm achieves higher ROIs than the PPO agent for certain altcoins. However, the average ROIs for all altcoins remain positive for both agents, with the averages being close with no statistical significance.  Moreover, the $CRN_{PPO}$ produces positive ROIs for three altcoins~(Binance Coin, Ethereum, and Litecoin), and the DDPG agent generates positive results for two altcoins~(Binance Coin and Ethereum).
In particular, as seen in Figure~\ref{fig:performance}, the $CRN_{DDPG}$ demonstrates the highest annual ROI for Binance Coin among all models, at 8.30\%. Furthermore, the DDPG agent outperforms the PPO agent for Ethereum, with a 5.59\% annual ROI. However, the DDPG agent produces a negative annual ROI for Litecoin. In addition, both agents exhibit the same pattern for Ripple and Tether, generating negative ROIs; however, the $CRN_{DDPG}$ agent mitigates losses to a greater extent compared to the other agent.

Table~\ref{tab:preds} shows the Buy-and-Hold strategy underperforms for most altcoins.  
New entrants often speculate on the cryptocurrency market, hoping for substantial gains over time. However, as these results show, employing the Buy-and-Hold strategy is generally ineffective for most altcoins. This is particularly evident as the market capitalisation decreased significantly from the end of 2021 to April 2023. It is worth reiterating that the cryptocurrency market, during this period, experienced highly volatile conditions, dropping by almost 50\% (around 1.8 trillion US dollars), resulting in significant losses for many investors.

With the Buy-and-Hold strategy, Binance Coin is the only cryptocurrency to generate a positive annual ROI (3.44\%). This positive performance indicates Binance Coin's growth during the evaluation period. Conversely, the Buy-and-Hold strategy performs poorly with Ethereum, Litecoin, and Ripple, producing a negative annual ROI. Moreover, Tether maintains almost the same value, showing a slight positive annual ROI of 0.04\%. One reason is that Tether is typically pegged to a stable asset, which maintains its value despite market fluctuations. However, this characteristic also means achieving significant profit through Tether trading might be challenging due to its limited price movements.

\subsection{Analysing the Performance of CRN Framework}

To analyse the performance of the CRN framework, we compare two agents using the CRN and base RL approaches. The comparison results are presented in Table~\ref{tab:preds-Base_RL}. It is important to note that the RL base models only utilise OHLCV data for trading, and both models share the same reward function and algorithms.

\begin{table*}[h]
\Large
\caption{Comparison of ROI and Annual ROI for PPO and DDPG agents using CRN and Base RL approaches across five altcoins. Numbers in parentheses show the standard deviation. `Ann.' stands for `Annual.'}
\label{tab:preds-Base_RL}
\resizebox{\columnwidth}{!}{
\renewcommand{\arraystretch}{2}
\setlength{\tabcolsep}{4pt}
\begin{tabular}{@{}lrrrrrrrr@{}}
\cmidrule{2-9}
\textbf{} & \multicolumn{2}{c}{\textbf{CRN$_{\textbf{PPO}}$}} & \multicolumn{2}{c}{\textbf{Base RL$_{\textbf{PPO}}$}} & \multicolumn{2}{c}{\textbf{CRN$_{\textbf{DDPG}}$}} & \multicolumn{2}{c}{\textbf{Base RL$_{\textbf{DDPG}}$}} \\ 
\cmidrule{2-9}
 & \multicolumn{1}{c}{ROI} & \multicolumn{1}{c}{Ann.ROI} & \multicolumn{1}{c}{ROI} & \multicolumn{1}{c}{Ann.ROI} & \multicolumn{1}{c}{ROI} & \multicolumn{1}{c}{Ann.ROI} & \multicolumn{1}{c}{ROI} & \multicolumn{1}{c}{Ann.ROI} \\ 
\hline
\textbf{Binance Coin} & 12.93~(30.34)  & 7.35~(17.25)   & 4.63~(13.65) & 2.63~(7.76) & 14.59~(17.18) & 8.30~(9.77) & 5.63~(11.62) &  3.20~(6.61)  \\ 
\hline
\textbf{Ethereum} & 8.19~(44.56) & 4.65~(25.34) & 2.69~(4.77) & 1.53~(2.72) & 10.46~(40.15) & 5.59~(22.83) &5.39~(23.78)& 3.04~(13.52) \\ 
\hline
\textbf{Litecoin} & 5.13~(26.06) & 3.88~(19.70) & -6.41~(20.22) & -3.64~(11.49) & -0.74~(13.43) & -0.56~(10.15) &-4.89~(8.24) & -2.77~(4.68) \\ 
\hline
\textbf{Ripple} & -2.13~(15.80) & -1.21~(8.99) & -5.26~(17.45) & -2.98~(9.92) & -2.04~(36.24) & -1.16~(20.61) & -6.76~(9.69)&-3.84~(5.47) \\ 
\hline
\textbf{Tether} & -4.54~(0.18) & -2.58~(0.10) & -4.45~(0.14) & -2.53~(0.08) & -2.26~(1.88) & -1.28~(1.07) & -2.55~(2.11)& -1.44~(1.20) \\ 
\hline
\end{tabular}
}
\end{table*}

Overall, for the PPO agent, the CRN framework consistently outperforms the base RL model, except for Tether, where no substantial improvement is observed. Specifically, in the case of Binance Coin, the $CRN_{PPO}$ agent demonstrates a notably higher annual ROI of~(7.35\%) compared to the $Base RL_{PPO}$ agent, which achieves an annual ROI of~(2.63\%). A similar result is observed for Ethereum, with the $CRN_{PPO}$ agent outperforming the base PPO agent with an annual ROI of 1.53\%. For Litecoin, the $CRN_{PPO}$ agent substantially enhances trading performance. Although the RL agents could not achieve a positive balance at the end, the $CRN_{PPO}$ still achieves higher performance than the base RL model.
As explained, there is no change in performance for Tether, suggesting the different behaviour of Tether as an altcoin, which the framework is not able to capture accurately.

Furthermore, the DDPG agent with the CRN framework generally achieves better performance than the base RL model across most altcoins, with the exception of Litecoin and Tether, where improvements are minimal or negative. For Binance Coin, the $CRN_{DDPG}$ framework shows a substantial improvement, with an annual ROI of 8.30\%, compared to 3.20\% for the $Base RL_{DDPG}$ agent. Ethereum also experiences enhanced results under the CRN framework, achieving an annual ROI of 5.59\%, compared to the 3.04\% annual ROI of $Base RL_{DDPG}$ agent. Moreover, for Litecoin, both CRN and base RL strategies underperform; however, CRN still manages a slightly lesser loss. In the case of Ripple, while both agents result in a loss, the $CRN_{DDPG}$ agent's performance is less negative than the base model. Tether, similar to its behaviour in the PPO models, shows minimal to no improvement with the CRN framework.

\subsection{Analysing Decisions of CRN Agents}

Table~\ref{tab:algs_stats} and \ref{tab:algs_stats_DDPG} provide} detailed information about the decisions made by our $CRN_{PPO}$  and $CRN_{DDPG}$ agents for each altcoin. These statistics provide an overview of the behaviour of the agents to explore the decision-making process of them. The `Number of Actions' column indicates the frequency of the agent's buy, sell, and hold actions. Moreover, the `Average Position Size' column provides the average size of positions that the agent takes for buy and sell actions. It is worth reiterating that according to our strategy, the position size is between 40\% and 60\%. However, the position size can increase to 75\% in strong Up or strong Down} scenarios. 
The tables also present the DBN-generated predictions for each altcoin during the test period. The column labelled `Up' shows the frequency in which the DBNs predict a price increase for the following day. On the other hand, the `Down' column indicates the percentage of instances in which DBNs predict a daily price decrease.

\begin{table*}[h]
\centering
\caption{\textbf{\bm{$CRN_{PPO}$} Decision Analysis}~--~The table provides information on the frequencies of trade types and the average position sizes for trades executed by the agent with the utilisation of the PPO algorithm.
}
 \vspace{-0.1cm}
\label{tab:algs_stats}
\vspace{-0.2cm}
  \resizebox{\columnwidth}{!}{%
 \renewcommand{\arraystretch}{1.35}
\begin{tabular}{lccccccr}
\cmidrule{2-8}
\multicolumn{1}{c}{} & \multicolumn{3}{c}{\textbf{Number of Actions~(\%)}} & \multicolumn{2}{c}{\textbf{Average Position Size~(\%) }} & \multicolumn{2}{c}{\textbf{DBN Predictions~(\%)}}  \\ \cmidrule{2-8}
\multicolumn{1}{c}{} & \textit{\textbf{Buy}} & \textit{\textbf{Sell}} & \textit{\textbf{Hold}} &\textit{\textbf{Buy}} & \textit{\textbf{Sell}} & \textit{\textbf{Up}} & \textit{\textbf{Down}}    \\ \hline
\textbf{Binance Coin} & 51.56 & 16.25  & 32.19  & 61.17 & 58.73 & 54.39 & 45.61  \\ \hline
\textbf{Ethereum} &47.03   & 17.82 & 35.21 & 64.45 & 63.33 &52.90 & 47.10 \\ \hline
\textbf{Litecoin} & 29.73 & 30.77 &39.50 & 53.09 & 53.17 &53.81 & 46.19\\ \hline
\textbf{Ripple} & 30.27 & 27.46 & 42.28 & 61.12 & 60.58 &  50.16 & 49.84  \\ \hline
\textbf{Tether} & 31.30 & 28.95 & 39.75 & 52.88 & 53.22 &40.13 & 59.87 \\ \bottomrule
\end{tabular}
  }
\end{table*}

As can be seen in Table~\ref{tab:algs_stats}, for Binance Coin and Ethereum, the $CRN_{PPO}$ agent demonstrates similar statistical behaviour. The agent mainly takes buy actions for these altcoins, accounting for 51.56\% and 47.03\% of its total actions, respectively. On the other hand, hold actions accounted for 32.19\% and 35.21\% of the agent's actions for Binance Coin and Ethereum. This indicates that the agent chooses to hold onto these altcoins without making any trading decisions. Sell actions have a smaller percentage, 16.25\% for Binance Coin and 17.82\% for Ethereum. This suggests that the agent is relatively cautious in selling these altcoins, potentially waiting for more favourable price movements or market conditions before making decisions. Regarding the average position sizes, the PPO agent allocates a slightly higher percentage of available funds for buying compared to selling for both Binance Coin and Ethereum. The average position sizes for buy actions are 61.17\% for Binance Coin and 64.45\% for Ethereum, while the average position sizes for sell actions are 58.73\% for Binance Coin and 63.33\% for Ethereum. It indicates that the agent tends to invest a more significant portion of its available balance in these particular assets, indicating higher confidence in the profitability of buy actions.

Our agent demonstrates a relatively balanced allocation of buy and sell actions for Litecoin, with 29.73\% and 30.77\%. This suggests that the RL agent is actively engaged in both buying and selling Litecoin based on its market analysis. However, our agent still tends to perform more hold actions for Litecoin with 39.50\%. Additionally, the average position size for buy and sell actions for Litecoin is very similar, with 53.09\% and 53.17\%, which indicates a more cautious trading approach.

\begin{figure*}[h]
  \centering
  \begin{subfigure}{.5\textwidth}
    \centering
\includegraphics[width=0.98\linewidth, height=0.98\textheight, keepaspectratio]{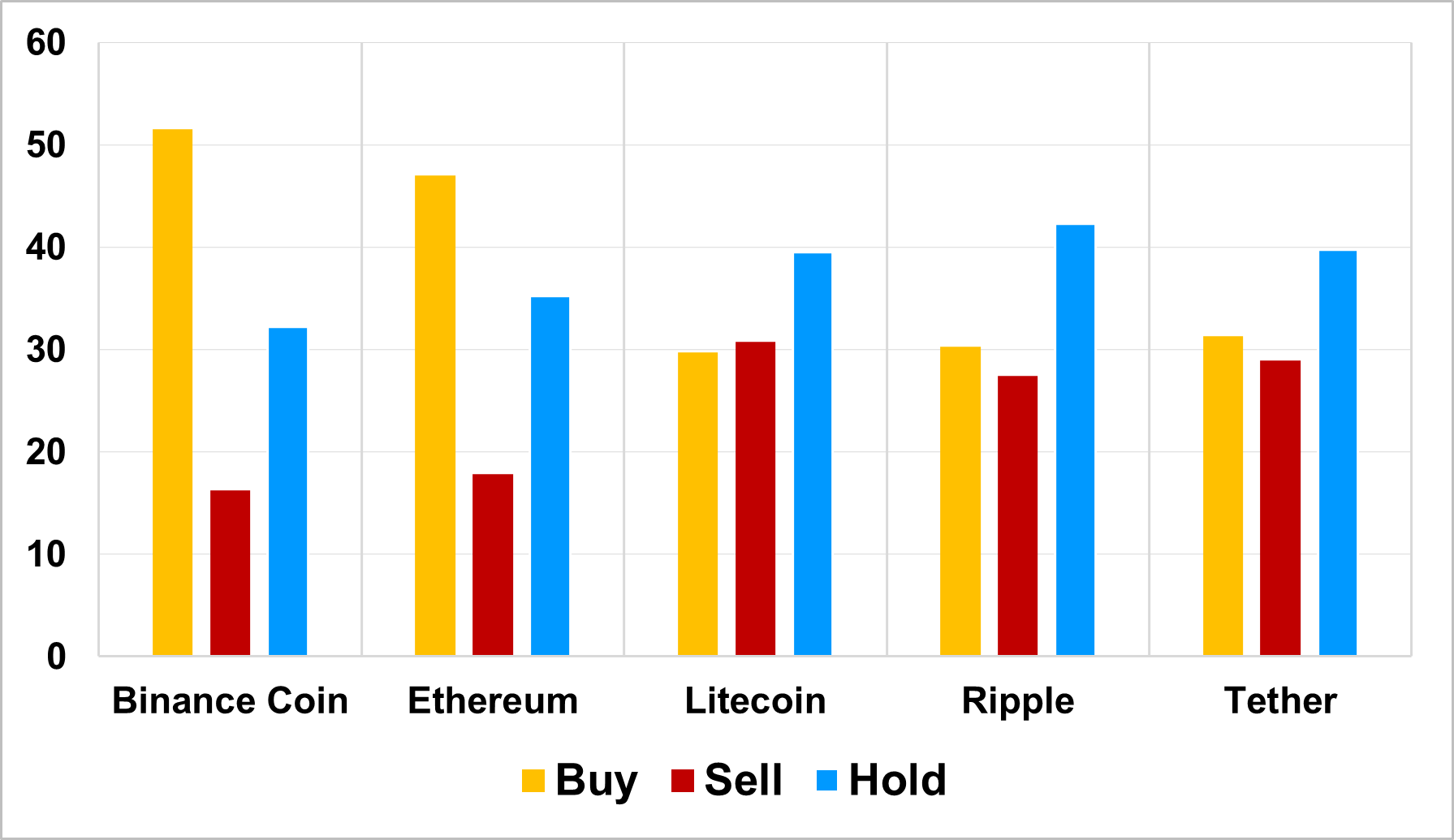}
    \caption{Number of Actions~(\%)}
    \label{subfig:Actions_PPO}
  \end{subfigure}%
  \begin{subfigure}{.5\textwidth}
    \centering   \includegraphics[width=0.98\linewidth, height=0.98\textheight, keepaspectratio]{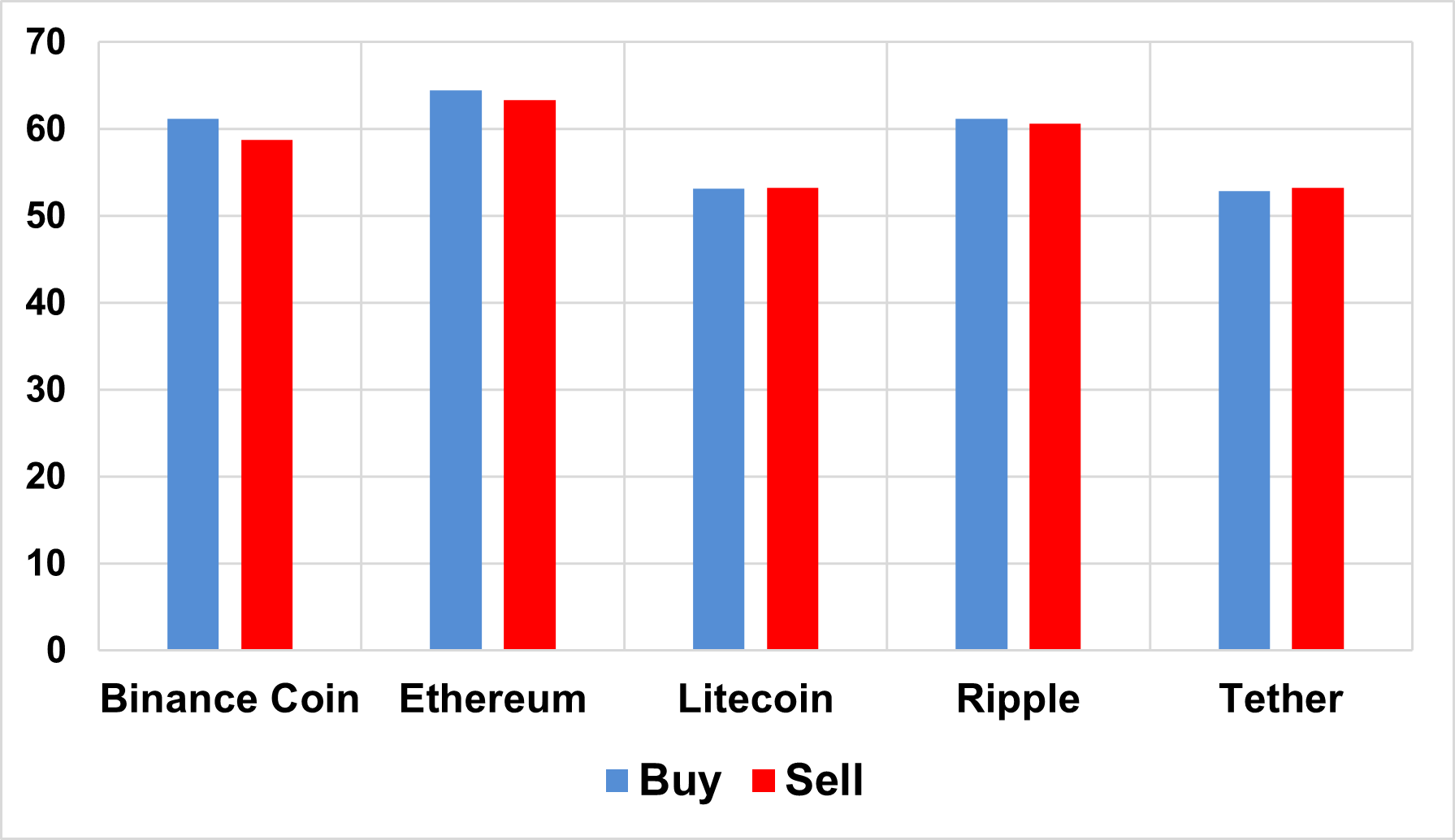}
    \caption{Average Position Size~(\%)}
\label{subfig:Position_Size_PPO}
  \end{subfigure}
  \caption{\textbf{\bm{$CRN_{PPO}$} Decision Visualisation} - These charts provide visualisation of the decisions made by the $CRN_{PPO}$ agent. Sub-figure(a) shows the percentage of each action taken relative to the total number of actions performed, while Sub-figure(b) illustrates the average position size for both sell and buy actions.}
  \label{fig:PPO_Decisions}
\end{figure*}

The agent's behaviour demonstrates a preference for holding positions for Ripple. Our agent has the highest proportion of hold actions~(42.28\%) for Ripple among all altcoins. This indicates that the agent selects to maintain its investment in Ripple without actively buying or selling,  possibly because the agent considers that the market conditions are not suitable for buying or selling. Moreover, the average position size for buy and sell actions is relatively similar, with 61.12\% for buy actions and 60.58\% for sell actions. The equal position sizes suggest that the agent aims to maintain a balanced portfolio by considering a longer-term investment strategy.
Furthermore, Tether has the same pattern as Ripple. The buy and sell agent's actions are almost evenly distributed, with buy actions of 31.30\%, sell actions of 28.95\%, and hold actions having the highest percentage of performed actions (39.75\%). Furthermore, the average position size for buy and sell actions is also the same, with 52.88\% and 53.22\%. Figure~\ref{fig:PPO_Decisions} provides a visualisation of the decisions made by ${CRN_{PPO}}$. The figure displays bar charts representing the number of actions and the average position size for each altcoin, which facilitates the comparison of the results across different altcoins. There are two notable observations. First, it is evident from Sub-figure~\ref{subfig:Actions_PPO} that the sell position has the lowest percentage in most altcoins concerning the number of actions. Second, the average sizes of the buy and sell positions are very close for each altcoin (Sub-figure~\ref{subfig:Position_Size_PPO}.)

The comparison between the decisions of the RL agents and the DBNs' predictions can provide valuable insights into the trading strategies of the agents. It is important to note that the DBNs produce two predictions (Up and Down). However, the agent has the flexibility to choose among three actions (Buy, Sell, and Hold). This distinction makes the comparison more challenging; however, the analysis still reveals meaningful patterns for certain altcoins.

Considering the decisions of {$CRN_{PPO}$} in Table~\ref{tab:algs_stats}, it is notable that the agent's decisions tend to align with the predictions generated by the DBNs for Binance Coin and Ethereum. 
In particular, DBNs generate a higher frequency of Up predictions, signalling more potential buy opportunities, and correspondingly, the RL agent also executes a significantly greater number of buy actions. Moreover, as can be seen for Ripple, where the percentages of Up and Down predictions are nearly identical, indicating market volatility, the PPO agent also adopts a cautious strategy, holding its position rather than performing buy or sell actions.

\begin{table*}[h]
\centering
\caption{\textbf{\bm{$CRN_{DDPG}$} Decision Analysis}~--~The table provides information on the frequencies of trade types and the average position sizes for trades executed by the agent with the utilisation of the DDPG algorithm.
}
 \vspace{-0.1cm}
\label{tab:algs_stats_DDPG}
\vspace{-0.2cm}
  \resizebox{0.9\columnwidth}{!}{%
 \renewcommand{\arraystretch}{1.35}
\begin{tabular}{lccccccl}
\cmidrule{2-8}
\multicolumn{1}{c}{} & \multicolumn{3}{c}{\textbf{Number of Actions}} & \multicolumn{2}{c}{\textbf{Average Position Size}} & \multicolumn{2}{c}{\textbf{DBN Predictions}} \\ \cmidrule{2-8} 
\multicolumn{1}{c}{} & \textit{\textbf{Buy}} & \textit{\textbf{Sell}} & \textit{\textbf{Hold}} &\textit{\textbf{Buy}} & \textit{\textbf{Sell}} & \textit{\textbf{Up}} & \textit{\textbf{Down}}    \\ \hline
\textbf{Binance Coin} & 24.43 & 26.21 &49.36  & 58.96 & 57.13  &54.39 & 45.61 \\ \hline
\textbf{Ethereum} & 26.12 & 25.30 & 48.58 & 61.37 & 61.01& 52.90 & 47.10 \\ \hline
\textbf{Litecoin} & 26.31 &24.56  & 49.13&50.45  &50.36& 53.81 & 46.19\\ \hline
\textbf{Ripple} & 23.53 & 24.56 &51.92  &59.53& 60.08&  50.16 & 49.84\\ \hline
\textbf{Tether} & 53.45 & 15.69 & 30.86 & 45.87 & 29.83& 40.13 & 59.87\\ \bottomrule
\end{tabular}
  }
\end{table*}

\begin{figure*}[h]
  \centering
  \begin{subfigure}{.5\textwidth}
    \centering
\includegraphics[width=0.98\linewidth, height=0.98\textheight, keepaspectratio]{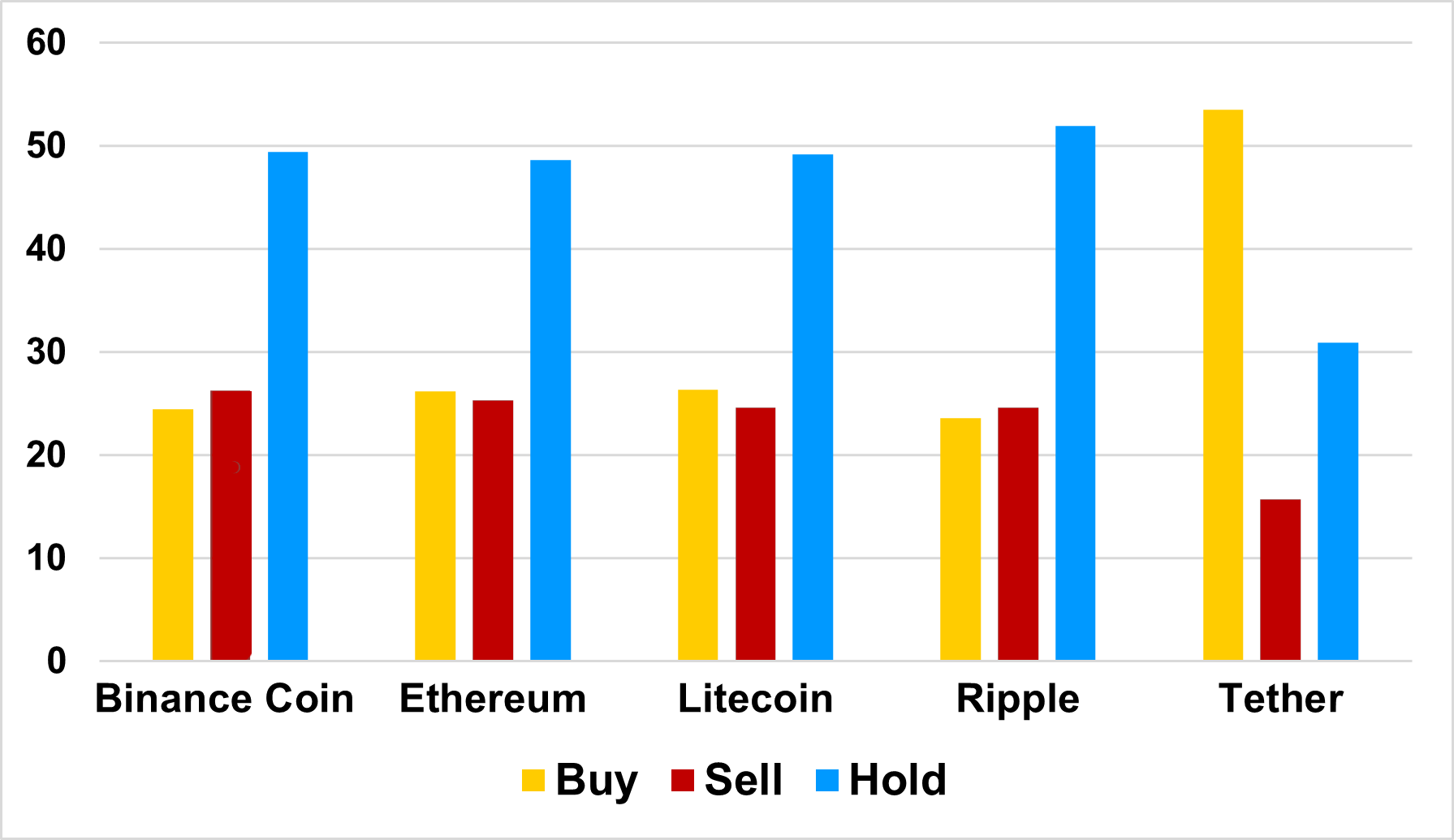}
    \caption{Number of Actions~(\%)}
    \label{subfig:Actions_DDPG}
  \end{subfigure}%
  \begin{subfigure}{.5\textwidth}
    \centering   \includegraphics[width=0.98\linewidth, height=0.98\textheight, keepaspectratio]{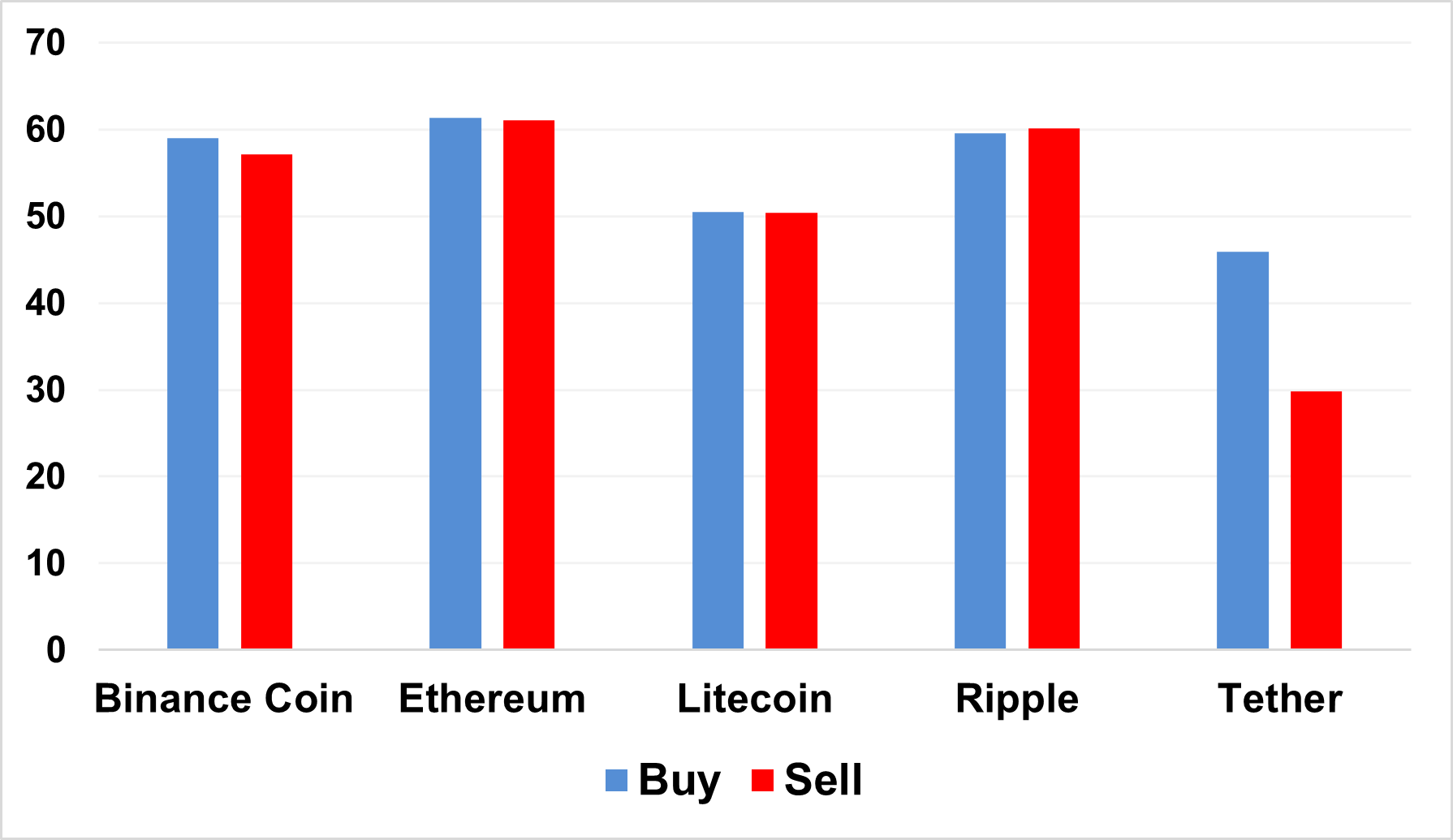}
    \caption{Average Position Size~(\%)}
\label{subfig:Position_Size_DDPG}
  \end{subfigure}
  \caption{\textbf{\bm{$CRN_{DDPG}$} Decision Visualisation} - These charts provide visualisation of the decisions made by the $CRN_{DDPG}$ agent. Sub-figure(a) shows the percentage of each action taken relative to the total number of actions performed, while Sub-figure(b) illustrates the average position size for both sell and buy actions.}
  \label{fig:DDPG_Decisions}
\end{figure*}

Regarding the decisions made by ${CRN_{DDPG}}$ agent presented in Table~\ref{tab:algs_stats_DDPG},  the hold action frequency is the highest among the three possible actions taken for Binance Coin, Ethereum, Litecoin, and Ripple~(around 50\%). 
Furthermore, the frequencies of buy and sell actions are nearly equal, suggesting a balanced trading approach for these cryptocurrencies. 
In contrast, the trading behaviour of the agent for Tether differs.
The agent performs more buy action with a frequency of 53.45\%, and the hold action is 30.86\%. Moreover, while  Binance Coin, Ethereum, Litecoin, and Ripple show relatively close average position sizes for both buying and selling, Tether has a different pattern. In particular, the average position size for buying Tether is noticeably larger at 45.87\%, compared to its selling position size, which is considerably smaller at 29.83\%. These differences between agents' decisions for Tether and others can be attributed to the fact that Tether is a stablecoin, while other altcoins are mining-based cryptocurrencies. 
 As observed, the agent adopts a distinctive trading strategy when interacting with Tether. These differences in agents' decisions regarding Tether can be attributed to its nature as a stablecoin, which is inherently different from other altcoins that are mining-based cryptocurrencies.
 Figure~\ref{fig:DDPG_Decisions} is provided to enhance understanding of the prior discussions regarding the decisions undertaken by the ${CRB_{DDPG}}$ agent. Sub-figure~\ref{subfig:Actions_DDPG} shows that the hold action clearly stands out as the dominant action for all altcoins except Tether. Additionally, the average percentage for buy and sell positions is almost the same for each, excluding Tether~(Sub-figure~\ref{subfig:Position_Size_DDPG}).

Comparing the ${CRN_{DDPG}}$ action analysis (Table~\ref{tab:algs_stats_DDPG}) and the predictions of the DBNs reveals an important observation that the agent is more conservative for all altcoins except Tether. In particular, while the DBNs generate more buy signals, the DDPG agent performs more hold action, which implies a conservative trading strategy that waits for stronger signals. On the other hand, despite a 59.87\% prediction of Down, this agent executes more buy actions (53.45\%) for Tether, suggesting it is employing a different approach for this stablecoin.

\section{Conclusions and Future Work}\label{Conclusions}
This study aims to develop an algorithmic trading system specifically designed for five popular altcoins, namely, Binance Coin, Ethereum, Litecoin, Ripple, and Tether.
We propose a framework named CRN, which serves as the foundational architecture of our trading system. The framework is designed to empower the RL agent through the integration of causal analysis to understand and model the probabilistic relationships among factors that can affect price movements. Utilising Bayesian networks, the framework identifies the most relevant features for each altcoin in the design of the RL agent's state observations. Furthermore, to enhance the agent's decision-making process and provide more accurate market information, we integrate dynamic Bayesian networks into our trading system and use its predictions. These predictions generate probabilistic buy and sell signals that guide the agent's trading decisions.
Moreover, we equip the RL agent with the capability to not only select trading actions but also determine the trade position sizes. Additionally, we create two RL agents based on different RL algorithms, Proximal Policy Optimisation
and Deep Deterministic Policy Gradient, to investigate the influence of selecting the RL algorithm on the performance of our trading system.
Furthermore, considering the high volatility and associated risks in the cryptocurrency market, we design our trading system with more conservative trading rules. This approach aims to mitigate potential risks and minimise exposure to large market fluctuations. In particular, we limit the position size of each trade to ensure that the agent does not take on excessive risk, thereby protecting our initial investment. 

Even though altcoins are part of the cryptocurrency market, each altcoin behaves differently and has its own specific ecosystem. Thus, one general conclusion cannot necessarily be drawn for all the altcoins together. Generally, both agents outperform the traditional benchmark model of the Buy-and-Hold. However, comparing the results from the agents based on two RL algorithms reveals similarities in terms of the return on investment as performance criteria. Despite varying performance levels, both RL agents perform well with Binance Coin and Ethereum, while they do not produce satisfactory results for Litecoin and Tether. The RL agents exhibit opposing results when trading with Ripple. Furthermore, we compare the performance of our framework to a baseline RL model to investigate the efficacy of the proposed framework. The comparison shows that the CRN consistently outperforms the baseline RL model. However, results indicate varying degrees of effectiveness across different cryptocurrencies. This comparative analysis highlights the potential advantages of the CRN framework and the integration of casual feature engineering into the RL framework, offering improved predictive power.
Additionally, analysing their decisions helps to understand the decision-making patterns of the agents. Generally, the analysis discloses a relatively balanced distribution between sell and buy positions for both agents. Meanwhile, the hold action occurs as their preferred trading strategy, which signals market uncertainty and an absence of clear buy or sell signals. 

The results and limitations observed in our study can be investigated in future research. One potential research direction is to explore higher trading frequencies such as hourly. Additionally, developing adaptable or parametric strategies to adjust the buy and sell lower and upper thresholds, including establishing the maximised threshold of the investment for buying or selling, dynamically can enhance the profitability of individual altcoins. Moreover, 
since the average ROIs for all altcoins of both agents in the analysis are positive,
incorporating a portfolio strategy with our existing RL trading system represents a promising direction for future research. By considering all altcoins as a single investment basket, we can potentially diversify risk, balance rewards, and enhance the overall trading performance of the system.

Furthermore, exploring a broader set of RL algorithms other than Proximal Policy Optimisation and Deep Deterministic Policy Gradient presents another direction for future research. Algorithms such as Trust Region Policy Optimisation and Soft Actor-Critic, which operate in a continuous action space, can be particularly beneficial to investigate. Another future research direction is to apply our proposed framework to high-frequency trading, where exploring high-frequency data could enhance the robustness of the framework and provide deeper insights into market dynamics. Finally, 
extending the use of the CRN framework to other domains requiring decision support systems, each with its unique input features, is another important research direction. This approach can offer deeper insights into the robustness of the framework and the feasibility of using it as a foundational architecture for designing RL-based systems.

\section*{Conflict of Interest}
The authors declare that they have no conflict of interest.

\section*{Appendix: Supplementary Data}
The data supporting the findings of this study are available in the Git repository at \url{https://github.com/bigrasam/CRNData.git}, providing open access to the research datasets for further investigation and replication\\

\newpage

\bibliographystyle{plainnat}  
\bibliography{cas-refs}

\newpage

\end{document}